\documentclass[conference]{IEEEtran}
\usepackage{array}
\usepackage{hyperref}
\IEEEoverridecommandlockouts
\usepackage{cite}
\usepackage{amsmath,amssymb,amsfonts}
\usepackage{algorithmic}
\usepackage{graphicx}
\usepackage{textcomp}
\usepackage{xcolor}
\def\BibTeX{{\rm B\kern-.05em{\sc i\kern-.025em b}\kern-.08em
    T\kern-.1667em\lower.7ex\hbox{E}\kern-.125emX}}

\makeatletter
\newcommand{\linebreakand}{%
  \end{@IEEEauthorhalign}
  \hfill\mbox{}\par
  \mbox{}\hfill\begin{@IEEEauthorhalign}
}
\makeatother

\begin{document}

\title{Converting Epics/Stories into Pseudocode using Transformers\\

}

\author{
\IEEEauthorblockN{Gaurav Kolhatkar}
\IEEEauthorblockA{
\textit{SCTR's Pune Institute of Computer Technology}\\
Pune, India \\
gauravk403@gmail.com}
\hspace{-0.5cm}
\and
\IEEEauthorblockN{Akshit Madan}
\IEEEauthorblockA{
\textit{SCTR's Pune Institute of Computer Technology}\\
Pune, India \\
akmadan17@gmail.com}
 \linebreakand
\IEEEauthorblockN{Nidhi Kowtal}
\IEEEauthorblockA{
\textit{SCTR's Pune Institute of Computer Technology}\\
Pune, India \\
kowtalnidhi@gmail.com}
\and
\IEEEauthorblockN{Satyajit Roy}
\IEEEauthorblockA{
\textit{SCTR's Pune Institute of Computer Technology}\\
Pune, India \\
satyajit12.roy@gmail.com}
 \linebreakand
\IEEEauthorblockN{ Sheetal Sonawane}
\IEEEauthorblockA{\textit{Associate Professor} \\
\textit{SCTR's Pune Institute of Computer Technology}\\
Pune, India \\
sssonawane@pict.edu}
}

\maketitle

\begin{abstract}
The conversion of user epics or stories into their appropriate representation in pseudocode or code is a time-consuming task, which can take up a large portion of the time in an industrial project. With this research paper, we aim to present a methodology to generate pseudocode from a given agile user story of small functionalities so as to reduce the overall time spent on the industrial project. Pseudocode is a programming language agnostic representation of the steps involved in a computer program, which can be easily converted into any programming language.  Leveraging the potential of Natural Language Processing, we want to simplify the development process in organizations that use the Agile Model of Software Development. We present a methodology to convert a problem described in the English language into pseudocode. This methodology divides the Text to Pseudocode conversion task into two stages or subtasks, each of which is treated like an individual machine translation task. Stage 1 is Text to Code Conversion and Stage 2 is Code to Pseudocode Conversion. We find that the CodeT5 model gives the best results in terms of BLEU score when trained separately on the two subtasks mentioned above. BLEU score is a metric that is used to measure the similarity between a machine-translated text and a set of reference translations.

\end{abstract}


\section{Introduction}
Efficiency of work is of the highest importance in modern organizations and businesses. A majority of the workplaces today use the Agile Model for software development. Agile is a software development approach based on iterative development, wherein tasks are divided into smaller iterations or sprints. In Agile project management tools such as Jira are used to document the user requirements in the form of epics or user stories. Developers need to understand these requirements and write code for the same. However, a significant amount of development time and efforts can be saved by automating the process of code/pseudo code generation, especially for simple or repetitive problems
that have been solved before. The motivation of our research paper is to simplify the work of developers so that they can focus on more complex tasks and in the process, to optimize the software development lifecycle. \newline

Jira is a software application used for issue tracking and project management. It is widely used by agile development teams to track bugs, stories, epics, and other tasks. Epics are large bodies of work that can be broken down into a number of smaller tasks
(called stories). Stories, also called “user stories,” are short requirements or requests written from the perspective of an end user. Our aim is to convert epics/stories to pseudo code. \newline

Despite the advantages of the Agile Model, developing software may still be a difficult and drawn-out process, especially when it comes to translating user requirements into functional code. Developers often translate user epics or stories manually into code as part of this process, which can take a lot of time and work. \newline

Our study intends to investigate the potential of utilising machine learning methods and natural language processing to automate the process of generating code and pseudocode from user stories in order to address this difficulty. By doing this, we hope to streamline developers' tasks, enhance the software development lifecycle, and boost the effectiveness of the entire industrial project. \newline

Recent advancements in the field of natural language processing have made it possible to automate a variety of formerly manual operations. Recent developments in deep learning, in particular, have made it possible to create sophisticated natural language models that can extract context and meaning from text input. By utilising these models, we can quickly and efficiently create code or pseudocode from user stories, relieving the pressure on developers. \newline

Our study will examine a variety of currently used methodologies, including as rule-based systems, statistical models, and deep learning techniques, for producing code or pseudocode from user stories. We will assess the benefits and drawbacks of each methodology and suggest a novel strategy that makes the most of their advantages.
\newline

Overall, our research paper aims to make software development more efficient and effective by automating the process of code/pseudocode generation from user stories. By doing so, we hope to free up developers to focus on more complex tasks, reduce the risk of errors, and ultimately deliver better software products to users.

\section{Literature Survey}

Saeki, Horai, and Enomoto's groundbreaking work in 1989 \cite{b15} marked the beginning of the path to close the gap between natural language and programming languages. With the aid of this fundamental discovery, software development could be sped up by converting natural language specifications into code. This work served as a conceptual starting point but also provided a framework for further investigation. \newline 

Cohn's important book "User Stories Applied" from 2004 \cite{b13} became a cornerstone as the software development industry turned towards more agile approaches. User stories are critical to Agile development, and Cohn underlined the value of quickly turning them into code. User stories have since taken on a key role in Agile workflows, signaling a dramatic turn in the field. \newline 

The emergence of automated code generation tools accelerated the shift from user stories written in plain language to programming code. Notably, Pseudogen was presented by Oda et al. in 2015 \cite{b19}. Even though it wasn't just concerned with text-to-pseudocode conversion, this work was a significant advancement in automating the conversion of code from descriptions written in plain language. \newline 

With the introduction of deep learning and neural networks, the conversion of text to pseudocode reached a turning point. In 2018, Devlin et al. unveiled BERT \cite{b1}, a revolutionary approach for interpreting natural language. Code generation underwent a revolution thanks to BERT's ability to record context, which made it possible to comprehend programming-related language more sophisticatedly. \newline 

By pre-training on data flow, GraphCodeBERT \cite{b2} expanded BERT's coding capabilities in 2021. This ground-breaking method, which made a substantial advancement in the discipline, used deep bidirectional transformers to improve code representation and comprehension. By addressing the need for models that could comprehend both the code and the data flow within it, generated pseudocode was of higher quality. \newline 

A unified approach for program understanding and citation generation was proposed by Ahmad et al. in 2021 \cite{b3}, signaling the convergence of various research strands. By providing a thorough understanding of program structures, this work aimed to improve code generation. It emphasized the possibility of improving software development through an understanding of the entire development cycle, from user stories to code generation. \newline

The value of standardized datasets became clear as the field developed. By serving as benchmarks for text-to-code and code-to-pseudocode conversion methods, datasets like the MBPP Dataset \cite{b26} and the ASE15 Django Dataset \cite{b27} significantly contributed to the advancement of technology. Researchers have been able to rigorously assess and compare their models thanks to these datasets. \newline 

Models like Code2Text \cite{b8} and DeepPseudo \cite{b10} have become more popular recently. These models improve the precision and applicability of text-to-pseudocode conversion by combining pre-training, deep learning, and code feature extraction. They offer reliable solutions for this task and are at the cutting edge of technology. These models are made to comprehend both the structure of the programming context and the code, allowing for more precise pseudocode generation. \newline

\section{Methodology}

\begin{figure*}[htbp]
  \centering
  \includegraphics[width=0.8\textwidth]{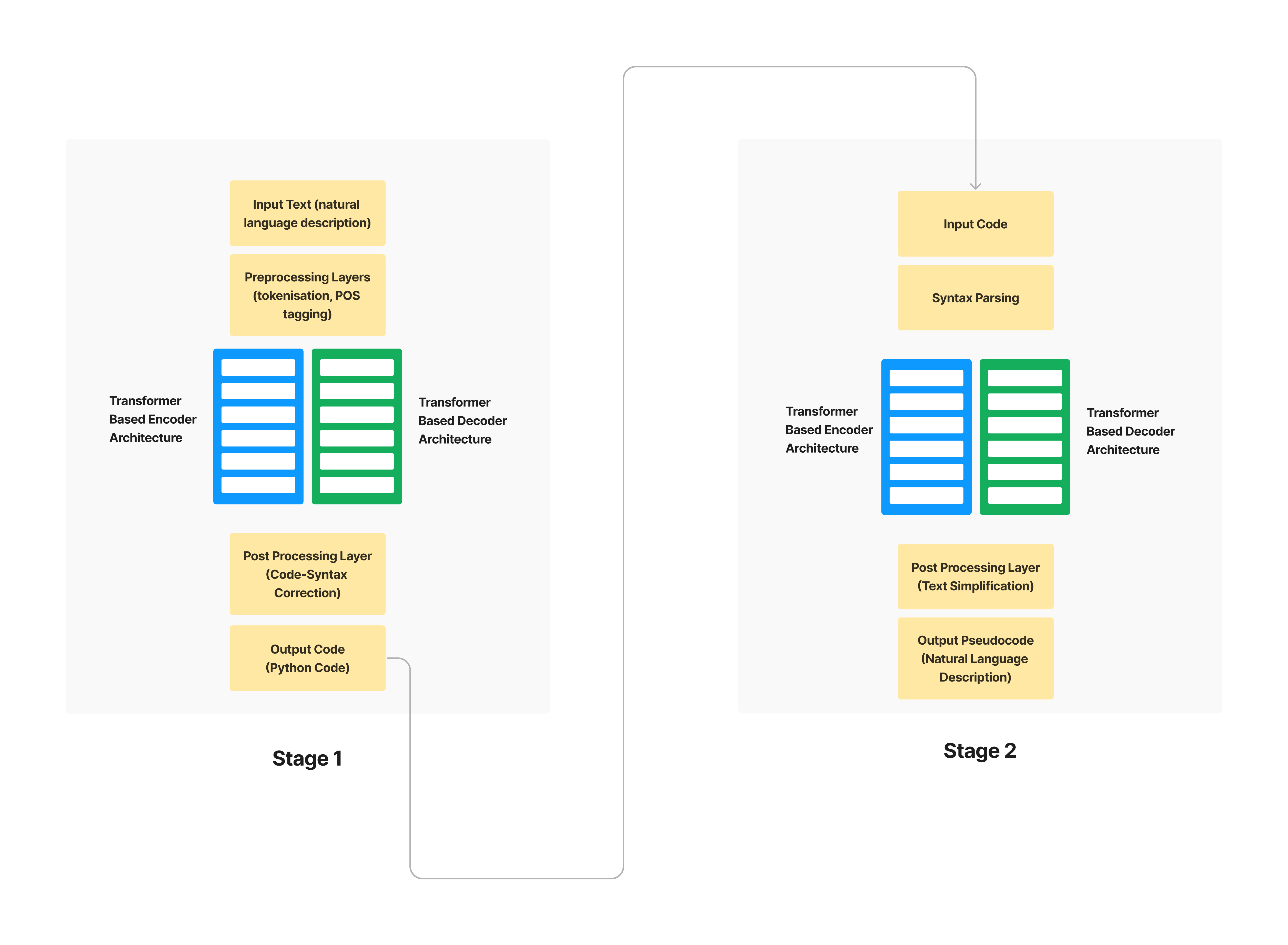}
  \caption{Stages of the proposed approach}
  \label{fig:stages_method}
\end{figure*}

\subsection{Stages of our proposed approach}

Our research paper aims to generate pseudocode from a given English language prompt. In order to achieve this task, we have divided the workflow into two stages, namely \hyperref[subsec:text-to-code]{Text to Code Conversion}  and \hyperref[subsec:code-to-pseudocode]{Code to Pseudocode Conversion}. We have fine-tuned \hyperref[subsec:code-t5]{CodeT5 Model} to get the required output. The performance of the model was measured in the form of \hyperref[tab:results]{BLEU score}.
\newline \newline \newline \newline

\subsubsection{Text to Code Conversion} 
\label{subsec:text-to-code}

This stage converts the initial Natural Language text input into Python code. Python was chosen as the intermediate representation language due to its common use and availability of datasets on Python code.
Text-to-code generation produces program code in a programming language from a natural language description of a program as its input. An encoder (such as a transformer-based design) creates a set of hidden states from the input text after preprocessing it (for example, tokenization and part-of-speech tagging). The output program code is produced by a decoder (such as an LSTM-based architecture) using these concealed states as input. The final result is then returned after postprocessing (for example, to fix syntax mistakes). \newline 
\begin{itemize}
\item  \textbf{Input Text (natural language description)}: 
\newline
This refers to the initial text input that is to be converted into code or pseudocode. This could be a sentence, a paragraph, or a longer piece of text that describes a computational task or problem to be solved.\newline

\item \textbf{Preprocessing Layers (tokenization, POS tagging)}:
\newline
This involves preparing the input text for further processing by breaking it down into individual words or symbols (tokenization) and identifying the part of speech of each token (POS tagging). Tokenization is the process of dividing text into meaningful units, or tokens, while POS tagging is the process of labeling each token with its corresponding part of speech, such as noun, verb, adjective, etc.\newline

\item \textbf{Encoder (transformer-based architecture)}:
\newline
The encoder takes in the preprocessed text and processes it using a transformer-based architecture, which learns to map the input text to a numerical representation that captures its meaning. The transformer-based architecture is a type of neural network architecture that has achieved state-of-the-art performance in many natural language processing tasks.\newline

\item \textbf{Decoder (transformer-based architecture)}:
\newline
The decoder takes in the encoded representation of the input text and generates the corresponding code or pseudocode output. The decoder also uses a transformer-based architecture and is trained to generate code or pseudocode that is consistent with the input text.\newline

\item \textbf{Postprocessing Layers (code syntax correction)}:
\newline
This stage involves checking the generated code for syntax errors and correcting them to ensure that it adheres to the syntax rules of the target programming language. This is an important step to ensure that the generated code is syntactically correct and can be executed without errors.\newline

\item \textbf{Output Code (programming language code)}:
\newline
This refers to the final output of the text-to-code generation process, which is a block of code written in a programming language. The programming language used for the output code depends on the task or problem being solved.\newline
\end{itemize}

\subsubsection{Code to Pseudocode Generation}
\label{subsec:code-to-pseudocode}
The Python code generated in stage one is converted to pseudocode form.
In code-to-pseudocode generation, a programming language's programme code serves as the input, and a simplified, high-level description of the code is produced as the output in everyday English. The input code is first preprocessed (for example, by parsing the syntax), and after that it is sent via an encoder (for example, a transformer-based architecture) to produce a set of hidden states. The decoder (also a transformer-based architecture) uses these hidden states as input and produces the output pseudocode. The final result is then postprocessed (for example, to make the text simpler) and returned as the output pseudocode. \newline 
Both stages of the proposed approach are analogous to a language translation task. We use an encoder-decoder transformer model to first convert the English text to Python code, and then the Python code to pseudocode.  \newline 
\begin{itemize}
\item \textbf{Preprocessing Layers (syntax parsing)}:
\newline
This stage involves preparing the input code for further processing by breaking it down into its component parts and identifying the relationships between those parts (syntax parsing). Syntax parsing is the process of analyzing the structure of the code and identifying its constituent parts, such as variables, functions, and control structures.\newline

\item \textbf{Encoder (transformer-based architecture)}:
\newline
The encoder takes in the parsed code and processes it using a transformer-based architecture to learn a numerical representation that captures its meaning. This encoder is similar to the one used in the text-to-code generation process, but it is trained on code rather than natural language text.\newline

\item \textbf{Decoder (transformer-based architecture)}:
\newline
The decoder takes in the encoded representation of the input code and generates the corresponding pseudocode output. This decoder is similar to the one used in the text-to-code generation process, but it is trained to generate pseudocode instead of code.\newline

\item \textbf{Postprocessing Layers (text simplification)}:
\newline
This stage involves simplifying the generated pseudocode to make it more easily understandable to humans. This may involve removing redundant or ambiguous information, simplifying complex expressions, or rephrasing complex statements in simpler terms.\newline

\item \textbf{Output Pseudocode (natural language description)}:
\newline
This refers to the final output of the code-to-pseudocode generation process, which is a simplified natural language description of the input code. The output pseudocode is designed to be more easily understandable to humans than the original code, and may involve rephrasing complex statements in simpler terms, removing redundant information, and simplifying complex expressions. The pseudocode may be used as a higher-level description of the code, making it easier to understand and maintain.
\end{itemize}

\begin{table}[h]
\centering
\caption{Datasets used}
\label{tab:dataset}
\renewcommand{\arraystretch}{1.5}
\begin{tabular}{|p{1.5cm}|p{2cm}|p{2cm}|p{1.5cm}|}
\hline
\textbf{Stages} & \textbf{Dataset} & \textbf{Programming Language} & \textbf{Number of Samples in the Dataset} \\ \hline
Text to Code           & MBPP            & Python            & 974            \\ \hline
Code to Pseudocode           & Django            & Python            & 16000            \\ \hline
\end{tabular}
\end{table}

\paragraph*{As shown in Table, our research utilized two datasets for the Text to Code and Code to Pseudocode stages. The first dataset, MBPP, consisted of 974 samples and was used for the Text to Code stage. The second dataset, Django, consisted of 16,000 samples and was used for the Code to Pseudocode stage. This table provides important information about the datasets used in our research, including their size and the programming language they were written in.}

\subsection{Transformers}

Transformers are attention-based models that don't use the typical recurrent layers found in encoder-decoder designs, but rather use multi-headed self-attention. Word embeddings from the input sequence are sent to the first encoder.

The data is then transformed and transmitted to the next encoder. The final encoder in the encoder-stack sends its output to every decoder in the stack of decoders. For translation tasks, the Transformer may be trained significantly more quickly than designs based on recurrent or convolutional layers.

The encoder and decoder layers in the Transformer architecture \ref{fig:t5_architecture} each employ multi-headed self-attention processes as opposed to conventional recurrent or convolutional layers. In order to produce a sequence of context embeddings, the encoder takes in a sequence of input embeddings and processes it through several layers. One token at a time, the decoder creates an output sequence using the context embeddings and a series of target embeddings. The model is tuned during training to reduce the discrepancy between the goal sequence and the anticipated output sequence. The Transformer is ideally suited for jobs that involve comprehending long-range dependencies because of its self-attention mechanism, which enables it to pay attention to various input and output sequences based on their relevance to the present prediction.


\subsection{CodeT5 Model}
\label{subsec:code-t5}
Modern neural language model CodeT5 was created with the goal of producing excellent source code from natural language inquiries. It is founded on the T5 architecture, which uses a framework for encoders and decoders based on transformers. The model is pre-trained using extensive corpora of natural language and code, which enables it to accurately capture the nuanced relationships between normal language and code.
An encoder that handles natural language inputs and a decoder that produces outputs in the form of code make up the architecture of CodeT5. Each token's contextual representation is produced by the encoder, a multi-layer transformer that analyses the incoming text. The decoder, which transforms the encoder's output into a series of code tokens, is also a multi-layer device. CodeT5 also employs a novel copy mechanism that allows it to directly copy tokens from the input text to the output code, improving the model's ability to handle rare or out-of-vocabulary words. We train CodeT5 for 40 epochs for the Text to Code conversion task, and for 5 epochs for the Code to Pseudocode conversion task.

\subsection {Rule Based Approach}
This method makes use of a Python script to convert python code given to the script as input in the form of a .py file.
The Python code given as input is converted into pseudocode with the help of a fixed set of rules. 
\newline
There are three types of rules:
\begin{itemize}
\item Basic conversion rules
\item Prefix conversion rules
\item Advanced conversion rules
\end{itemize}
The Python code is scanned line by line and pseudocode is generated for every corresponding set of code.
The output of the script is the pseudocode generated in the form of a .txt file 

\subsection{Disadvantages of Rule-based approach}

The proposed approach is more robust, flexible and dynamically adaptive as compared to the rule-based approach.
The scope of the rule-based approach is limited as compared to the extensive scope of our proposed approach.
 On human evaluation and making use of evaluation metrics like the BLEU score, the proposed approach performs better than the rule-based approach.
 The rule-based approach fails to generate appropriate pseudocode if keywords beyond the scope of the fixed set of conversion rules exist in the code. However, such a case can be appropriately handled by the proposed approach that makes use of transformers.
In summary, the proposed approach is more efficient and accurate as compared to the rule-based approach.

\section{Performance Analysis}
We have used BLEU score to analyse the performance of our Machine Learning Models.
\subsection{BLEU Score}
A popular technique for assessing the quality of machine-generated translations against a set of reference translations created by humans is called BLEU (Bilingual Evaluation Understudy) \cite{b28}. Its foundation is the comparison of the n-gram overlap between the machine translation and the reference translations. \newline 
The precision score, which assesses how many n-grams in the machine-generated translation also present in the reference translations, is first computed for each n-gram up to a specific maximum length (usually 4), before the BLEU score is determined. In order to discourage short and inadequate translations, the precision score is then adjusted by a brevity penalty factor that considers how long the machine-generated translation is in comparison to the reference translations. The final step is to integrate the updated accuracy scores into a geometric mean to create the overall BLEU score.

\begin{table}[h]
\centering
\caption{BLEU Score}
\label{tab:results}
\renewcommand{\arraystretch}{1.5}
\begin{tabular}{|p{2.5cm}|p{2.5cm}|p{2.5cm}|}
\hline
\textbf{Stages} & \textbf{Dataset Used} & \textbf{BLEU Score Value}   \\ \hline
Text to Code      & MBPP     & 0.4                   \\ \hline
Code to Pseudocode       & Django    & 0.74                     \\ \hline
\end{tabular}
\end{table}

The BLEU score values obtained in our study, as shown in Table, were 0.4 for Text to Code and 0.74 for Code to Pseudocode. The BLEU score for the first stage of the project denotes understandable to good translations, which is evident from the syntactical correctness of the generated Python code. However, there may be logical errors in the Python code due to insufficient training data. The BLEU score for the second stage points to high quality translations, with the generated pseudocode exhibiting a high level of logical and structural correctness. This clearly shows that, given sufficient training data, this methodology promises to give excellent results for the Text to Pseudocode conversion task.

\section{Conclusion}

This paper examines the enormous potential for programming work automation
offered by natural language processing. We propose a two-stage methodology to convert English language user stories into pseudocode. The advantage of pseudocode is that it allows easy conversion into any programming language of the developer’s choice. The two stages of this approach are text to code conversion and code to pseudocode conversion. Each of these stages is treated as a language translation task. We use the CodeT5 model for this task, getting a BLEU score of 0.4 for Stage 1 and 0.74 for Stage 2. Our proposed system simplifies the software development process in organizations.
\newline
In the future, we plan to curate a larger dataset for English text to Python code conversion. We can obtain higher accuracy and better generalisation to new examples with a larger and more varied dataset. This would necessitate considerable data gathering, but it might open the door to more efficient and useful text to code conversion methods. Additionally, a pertinent text to pseudocode dataset might be produced to help simplify the conversion architecture.

\end{document}